\documentclass[11pt]{article}
\usepackage{fullpage,graphicx,psfrag,amsmath,amsfonts,verbatim}
\usepackage{xcolor}
\usepackage{amsthm}
\usepackage[small,bf]{caption}
\usepackage{authblk}

\usepackage{amsmath}
\usepackage{amssymb}
\usepackage{mathtools}
\usepackage{amsthm}
\usepackage{amsfonts}
\usepackage{algorithm}
\usepackage{algorithmic}

\usepackage{microtype}
\usepackage{graphicx}
\usepackage{subfigure}
\usepackage{booktabs} % for professional tables
\usepackage{hyperref}

\allowdisplaybreaks

\title{TrackDiffuser: Nearly Model-Free Bayesian Filtering with Diffusion Model}
\author[1]{Yangguang He}
\author[2]{Wenhao Li\thanks{\texttt{whli@tongji.edu.cn}}}
\author[3]{Minzhe Li}
\author[4]{Juan Zhang}
\author[5]{Xiangfeng Wang}
\author[2,1]{Bo Jin\thanks{\texttt{bjin@tongji.edu.cn}}}

\affil[1]{Shanghai Research Institute for Intelligent Autonomous Systems, Tongji University}
\affil[2]{School of Computer Science and Technology, Tongji University}
\affil[3]{School of Aeronautics and Astronautics, Shanghai Jiaotong University}
\affil[4]{Shanghai Zhangjiang Institute of Mathematics}
\affil[5]{School of Computer Science and Technology, East China Normal University}

\date{}
\begin{document}
\maketitle

\begin{abstract}
% State estimation problems have significant application insights in various areas and Bayesian filtering is a common solution to the problem. 
% Classical model-based Bayesian filtering cannot deal with inaccurate state space model (SSM) and the lack of prior domain knowledge that often occurs in practice.
% In this paper, we present TrackDiffuser, a generative and nearly model-free Bayesian filtering.
% It can obtains the posterior distribution of the system state through the unique denoising mechanism of the diffusion model. 
% This novel progressive denoising diffusion strategy can implicitly learn system dynamics from the data, enabling it mitigate the effects of inaccurate SSM.
% By modeling Bayesian filtering as a measurement-conditional diffusion model, we demonstrate how we can circumvent the need for prior system domain knowledge including measurements model and noise characteristics. 
% To retain the interpretability of the model-based method, we design an implicit predict and update filtering framework combined with diffusion model. 
% Numerical experiments show that Bayesian filtering with diffusion model is an effective state estimation solution, in challenging scenarios like non-linear measurement and non-Gaussian noise, and still outperforms classical and data-driven methods. 
% It can still handle inaccurate SSM scenarios, exhibiting certain robustness.

State estimation remains a fundamental challenge across numerous domains, from autonomous driving, aircraft tracking to quantum system control. 
Although Bayesian filtering has been the cornerstone solution, its classical model-based paradigm faces two major limitations: 
it struggles with inaccurate state space model (SSM) and requires extensive prior knowledge of noise characteristics.
We present TrackDiffuser, a generative framework addressing both challenges by reformulating Bayesian filtering as a conditional diffusion model. 
Our approach implicitly learns system dynamics from data to mitigate the effects of inaccurate SSM, while simultaneously circumventing the need for explicit measurement models and noise priors by establishing a direct relationship between measurements and states. 
Through an implicit predict-and-update mechanism, TrackDiffuser preserves the interpretability advantage of traditional model-based filtering methods. 
Extensive experiments demonstrate that our framework substantially outperforms both classical and contemporary hybrid methods, especially in challenging non-linear scenarios involving non-Gaussian noises. 
Notably, TrackDiffuser exhibits remarkable robustness to SSM inaccuracies, offering a practical solution for real-world state estimation problems where perfect models and prior knowledge are unavailable.
\end{abstract}

\section{Introduction}

State estimation, the task of inferring hidden system states from noisy observations, is fundamental to numerous applications ranging from autonomous systems to economic forecasting \cite{8961145, Kaufmann2023, athans1974importance, de2020normalizing}.
The estimation process integrates a state-space model (SSM) including a state-evolution model, a measurement model, and noise characteristics within a Bayesian filtering framework to recursively estimate the posterior distribution of system states.

% While the Kalman filter (KF) provides the minimum variance optimal solution under the assumption of a Gaussian linear state-space model (SSM), many real-world problems are governed by non-linear dynamic systems. 
% Consequently, classical model-based (MB) filters have been proposed, including the Extended Kalman Filter (EKF) \cite{4982682, 1098671}, the Unscented Kalman Filter (UKF) \cite{Julier1997NewEO}, and the Particle Filter (PF) \cite{DELMORAL1997653, doi:10.1049/ip-f-2.1993.0015, 7079001}, to mitigate the complex dymanic challenge.

% However, The performance of MB methods is limited by the accuracy of the SSM \cite{9733186}. 
% Inaccurate prior information such as SSM mismatch may result in divergence. 
% However, in real-world complex application scenarios, systems are often non-cooperative, meaning that accurate SSM is frequently unavailable. 
% For instance, the system state-evolution model may not match reality, and even prior domain knowledge, such as noise statistics, may be unknown.
% In this scenario, such MB methods will fail to obtain an accurate posterior probability distribution. 

While the classical Kalman filter provides a minimum-variance optimal solution for linear Gaussian systems, real-world applications typically involve non-linear dynamics.
This prompts various model-based (MB) approaches such as Extended Kalman Filter (EKF) \cite{4982682, 1098671}, Unscented Kalman Filter (UKF) \cite{Julier1997NewEO}, and Particle Filter (PF) \cite{DELMORAL1997653, 7079001}. 
However, these methods face critical limitations: performance degradation in highly non-linear regions, sensitivity to model mismatch, and reliance on accurate prior knowledge of the system dynamics model and noise characteristics, when these factors are unavailable in practical scenarios \cite{9733186}.

Recent advances in simulation technology and sensor deployment \cite{10508326} have led to data-driven alternatives using deep learning, particularly RNNs \cite{6795963, 69e088c8129341ac89810907fe6b1bfe}, demonstrating remarkable capability in learning complex dynamics \cite{YAN2024102580, 10502278, 10320373}.
These architectures have been successfully integrated into Bayesian filtering frameworks \cite{9146931, 9761897, 9110734}, offering a promising solution to scenarios where accurate SSM or prior knowledge is unavailable. 
However, this end-to-end approach abandons valuable dynamic model structures, requiring substantially more training data and struggling with long-term dependencies and system variations. 
Hybrid approaches \cite{9733186, 10566495, 10120968, 10632588} attempt to balance model interpretability and data adaptability by maintaining the filtering framework while learning specific components. 
However, these solutions inherit the limitations of EKF through local linearization and Gaussian assumptions, and cannot satisfy the need for more robust and flexible approaches.

% Considering non-cooperative scenarios, we have limited prior domain knowledge of the non-linear SSM. Specifically, we only possess prior knowledge of the system dynamics, with no information about the noise model or even the measurement model.
% Our problem arises can we find a more general method to approximate the posterior probability density distribution?

The fundamental challenge lies in finding a unified framework, that can efficiently bridge the gap between physical principles and data-driven adaptation, particularly for highly non-linear systems where traditional linearization techniques and Gaussian assumptions break down. 
This motivates us to develop a flexible state estimator that preserves physical model structures while adapting to unknown system components through a novel probabilistic approach.

% In recent years, conditional generative models have produced impressive results across various fields \cite{ho2022video, xu2023versatile, anonymous2024physicsinformed, ajay2023is, 10.5555/3666122.3667354, fan2024mgtsd}. 
% Diffusion models have demonstrated strong capabilities in learning data distributions, e.g., trajectory data in offline RL \cite{ajay2023is}, time series data in time series forecasting \cite{10.5555/3666122.3667354}, motivating us to apply diffusion models to approximate the posterior probability density distribution in Bayesian filtering.

We propose TrackDiffuser, leveraging the diffusion model that has demonstrated remarkable capabilities in learning complex data distributions \cite{ho2022video, xu2023versatile, anonymous2024physicsinformed, ajay2023is, 10.5555/3666122.3667354, fan2024mgtsd}. 
Its progressive denoising process naturally decomposes complex distributions into a sequence of simpler transformations, making it ideal for approximating the posterior distribution in Bayesian filtering, especially for strongly non-linear systems where traditional assumptions fail.

% We propose a novel framework for Bayesian filtering, named TrackDiffuser that obtains the posterior distribution of the system state through the unique denoising mechanism of the diffusion model. 
% During training, in order to mitigate the effects of inaccurate SSM, we learn trajectories containing combined measurement sequences and state truths of real system dynamics.
% This novel progressive denoising diffusion strategy can implicitly learn system dynamics from the data, enabling it to mitigate the effects of inaccurate SSM. 
% By modeling Bayesian filtering as a measurement-conditional diffusion model, we establish a direct relationship between the measurement and the state, bypassing the need for prior system domain knowledge including the measurement model and the noise model.
% In inference, to retain the interpretability of the MB method, we integrate the Bayesian filtering framework into the generation process. 
% To ensure that the generated trajectory satisfies the system dynamics constraints, we perform the predict step by using the prior state estimate as the mean of the noise sampled during the denoising process. 
% We then perform the update step by recursively applying denoising using measurements-conditional guidance.

Unlike hybrid EKF-based methods that struggle with linearization and Gaussian assumptions, or purely data-driven approaches that abandon physical mechanisms, TrackDiffuser maintains model interpretability and adaptive capability to non-linear and non-Gaussian scenarios. 
%by measurement-conditional diffusion modeling. 
By modeling Bayesian filtering as a measurement-conditional diffusion model, TrackDiffuser establishes a direct relationship between measurements and states, that naturally handles non-Gaussian uncertainties in system dynamics and measurement processes. 
%Moreover, our framework naturally handles non-Gaussian uncertainties while preserving physical interpretability by integrating Bayesian filtering principles:
Moreover, TrackDiffuser preserves interpretability by integrating Bayesian filtering principles:
the predict step utilizes prior state estimates during denoising, while the update step recursively applies measurement-conditional guidance.

% We evaluate the performance of TrackDiffuser on a non-linear SSM under multiple experimental setups: linear and non-linear measurement scenarios, Gaussian and non-Gaussian noise, mismatched conditions, and real-world applications using the Michigan NCLT dataset \cite{10.1177/0278364915614638}. 
% Our TrackDiffuser outperforms KalmanNet and MB EKF, UKF, and PF.
Experiments across diverse scenarios, including non-linear dynamics, non-Gaussian noises, mismatched conditions, and a real-world long-time dataset, demonstrate that TrackDiffuser consistently outperforms both traditional model-based approaches and hybrid learning-based methods, such as KalmanNet, achieving a remarkable balance between physical principles and data-driven adaptation.

Our main contributions are as follows:
1) We pioneer the first diffusion-based Bayesian filtering framework, that fundamentally resolves the long-standing trade-off between model interpretability and handling uncertainty in state estimation;
2) The state estimation is formulated as a measurement-conditional denoising process, enabling direct modeling of non-linear dynamics and non-Gaussian uncertainties without relying on traditional approximations;
3) Consistent performance improvements are demonstrated across challenging scenarios, while requiring less prior domain knowledge than typical filters, demonstrating the effectiveness of physics-guided denoising in practical applications.

%DNN并不以原则性的方式并入诸如结构化SS模型之类的领域知识。因此，这些DD方法需要许多可训练参数和大型数据集，即使对于简单序列[23]也是如此，并且缺乏MB方法的可解释性。
%尽管设计建议分布努力很多，PF仍然可能无法近似后验分布

%contribution 
%1）提出一种新的计算后验分布的方式
%2）可以不需要知道h、w、v的信息完成推理，并且在SSM不匹配的情况下鲁棒性更好
%一种基于diffusion model新型的贝叶斯滤波器。通过直接学习后验概率分布的方式，规避掉对SSM模型中观测方程、噪声统计量的先验假设的依赖性，从而解决了对非线性系统中特定问题（噪声假设）定制不同解决方案的需求。通过在非线性模型（洛伦兹吸引）下开展广泛实验，说明TrackDiffuser具备从数据中学习状态后验概率分布的能力，我们还使用密歇根NCLT数据集进行定位证明了现实应用的能力。TrackDiffuser在非线性显著和SSM不匹配场景下优于KalmanNet以及MB EKF, UKF, and PF

% test\cite{10.5555/3666122.3667354}

\section{Related Work}

\subsection{Bayesian filtering}

Depending on whether deep learning is used, existing Bayesian filtering algorithms can be categorized into two groups, i.e., model-based and data-driven approaches.
Model-based approaches are based on the state space model. The most representative one is the EKF \cite {4982682, 1098671}, an extension of KF, which is based on local linearization with low computational efforts.
When the degree of nonlinearity is higher, the Unscented Transform based UKF \cite {Julier1997NewEO} is proposed, which computes the sigma points avoiding the computation of the Jacobi matrix.
For more complex SSMs, particularly a non-Gaussian process, Sequential Monte Carlo sampling methods have been proposed, such as the PF family  \cite{DELMORAL1997653, 7079001}, which distributes a large number of particles to approximate the posterior distribution.
However, filtering performances of these MB methods heavily depend on the accuracy of the SSM.

Alternatively, recent advances in simulation technology and sensor deployment \cite{10508326} have led to data-driven using deep learning. 
The current approaches are categorized into two types depending on whether a priori SSM is used or not.
The purely data-driven approach, i.e. end-to-end approach \cite{10320373, YAN2024102580}, abandons valuable dynamic model structures, and approximates the state-evolution function directly. 
However, even for simple sequences \cite{zaheer2017latent}, purely data-driven methods require vast amounts of data while lacking interpretability compared with MB methods.
The other is a hybrid approach combining SSM and RNN, which employs neural networks to dynamically adjust some of the parameters in SSM. 
A typical method is KalmanNet \cite {9733186}, which uses RNNs to dynamically adjust process and measurement noise parameters, thereby learning the optimal Kalman gain. 
Later, \cite{ 10288023} learns the optimal Kalman gain along with errors in IMU measurements. \cite{10566495} extended it to unsupervised learning, in which the Kalman gain is learned based on the Measurement Residual Log-Likelihood. 
Currently, hybrid methods dynamically adjust noise parameters, which can mitigate the impact of the SSM mismatch.

\subsection{Diffusion model}

Diffusion model has initially demonstrated strong capabilities in learning generative models for image and text data \cite{ho2022video, xu2023versatile}, and have since been applied in various domains for generative tasks. 
Recent work have used diffusion models to parameterize policies in offline reinforcement learning (RL), where \cite{ajay2023is, 10.5555/3618408.3619493} employed conditional diffusion models to generate trajectories aligned with high-reward conditions. 
In time series forecasting \cite{10.5555/3666122.3667354, fan2024mgtsd}, forecasters are designed as observation self-guidance diffusion models, iteratively refining the basic predictor forecasts by leveraging learned data distributions \cite{10.5555/3666122.3667354}.

% Diffusion models initially demonstrated strong capabilities in learning generative models for image and text data \cite{ho2022video, xu2023versatile}, they are now being applied across various domains for generative tasks. Recent work has used diffusion models to parameterize policies in offline reinforcement learning (RL), where \cite{ajay2023is, 10.5555/3618408.3619493} employed conditional diffusion models to generate trajectories aligned with high-reward conditions. In time series forecasting \cite{10.5555/3666122.3667354, fan2024mgtsd}, \cite{10.5555/3666122.3667354} designed forecasters as observation self-guidance diffusion models, iteratively refining the basic predictor forecasts using learned data distributions.

%tip最后一段可以再加一些扩散模型本身技术相关

\section{Preliminaries}

\subsection{State Space Model}

The state space model is fundamental to dynamic systems, and widely used in signal processing, control, and tracking. 
Consider a class of discrete-time non-linear dynamic system characterized by an SSM \cite{bar2004estimation}:
\begin{subequations}
\begin{align}
    {x}_t &= \mathbf{f}({x}_{t-1}) + \mathbf{w}_t \tag{1a}, \label{2.1.1} \\
    {z}_t &= \mathbf{h}({x}_t) + \mathbf{v}_t \tag{1b}, \label{2.1.2}
\end{align}
\end{subequations}
where $ {x}_t \in \mathbb{R}^{n_x} $ and \( {z}_t \in \mathbb{R}^{n_z} \) is the system state and measurement.
%at discrete time \( t \),
\( \mathbf{f}: \mathbb{R}^{n_x} \times \mathbb{R}^{n_w} \to \mathbb{R}^{n_x} \) and \( \mathbf{h}: \mathbb{R}^{n_x} \times \mathbb{R}^{n_v} \to \mathbb{R}^{n_z} \) denote the system motion and measurement models, respectively. 
The process noise \( \mathbf{w}_t \in \mathbb{R}^{n_v} \) and the measurement noise \( \mathbf{v}_t \in \mathbb{R}^{n_v} \) are uncorrelated, and have nominal covariance matrices \( \mathbf{Q} \in \mathbb{R}^{n_x \times n_x} \) and \( \mathbf{R} \in \mathbb{R}^{n_z \times n_z} \) under the Gaussian assumption.

\subsection{Bayesian Filter}

% QQQ
%Bayesian filtering estimates the state of a system recursively by updating the prior distribution based on given measurements and the system dynamic model. 
Bayesian filtering recursively estimates system states by updating the posterior distribution based on the system dynamics and measurements \cite{1228524}. The process involves two main steps:

\begin{equation}\label{2.2.1}
\textsf{Predict step}: p({x}_t|{z}_{1:t-1}) = \int p({x}_t|{x}_{t-1}) p({x}_{t-1}|{z}_{1:t-1}) d{x}_{t-1},
\end{equation}

\begin{equation}\label{2.2.2}
\textsf{Update step}: p({x}_t | {z}_{1:t}) = \frac{p({z}_t | {x}_t) p({x}_t | {z}_{1:t-1})}{p({z}_t | {z}_{1:t-1})},
\end{equation} where the predict step uses the system dynamic model to project the state estimate, 
and the update step incorporates new measurements to refine the posterior state estimate. 
However, directly applying Bayesian filtering in practice suffers from significant challenges.
%One major issue is the complexity of the state-evolution distribution \( p(x_t | x_{t-1}) \) and the measurement distribution \( p(z_t | x_t) \). 
In real-world scenarios, the state-evolution distribution \( p(x_t | x_{t-1}) \) and the measurement distribution \( p(z_t | x_t) \) are non-linear or involve unknown distributions, making the integrals in the predict step intractable. 
Additionally, the update step requires the computation of the normalizing constant \( p(z_t | z_{1:t-1}) \), which often involves high-dimensional integration, further complicating the process. 
These challenges necessitate the use of approximate methods, such as the EKF, UKF, and PF, to make Bayesian filtering computationally feasible.
% PF没有引用 

\subsection{Diffusion Probabilistic Models}

When the assumed SSM does not match the real-world data, the performance of model-based (MB) filters degrades significantly \cite{9733186}, making it impossible to obtain the posterior distribution.
% Here, we introduce the diffusion model, which is highly promising to learn data distributions.
Diffusion model is designed to learn the data distribution generatively from a given dataset $\mathcal{D}=\left\{{{\tau}}^i\right\}_1^N$. 
It approximate the data distribution by progressively adding noise to the data and learning how to denoise it in reverse.

The objective is to minimize the mean squared error between the true noise \( \epsilon \) and the noise predicted by the model. 
It is important to note that, the noise is unrelated to the noise in SSM, but rather a distinctive mechanism of the diffusion model. 
A simplified loss is $L = \mathbb{E}_{k, \tau_0, \epsilon} \left[ \| \epsilon - \epsilon_\theta(\tau_k, k) \|^2 \right]$ \cite{10.5555/3495724.3496298}, where \( \epsilon \sim \mathcal{N}(0, \mathbf{I}) \) is the noise sampled from a Gaussian distribution.
Through learning the reverse process, diffusion models can approximate complex posterior distributions, and provide a flexible alternative to traditional MB Bayesian filters.

\section{Method}

\begin{figure*}[htb!]
    \centering
    \includegraphics[width=\linewidth]{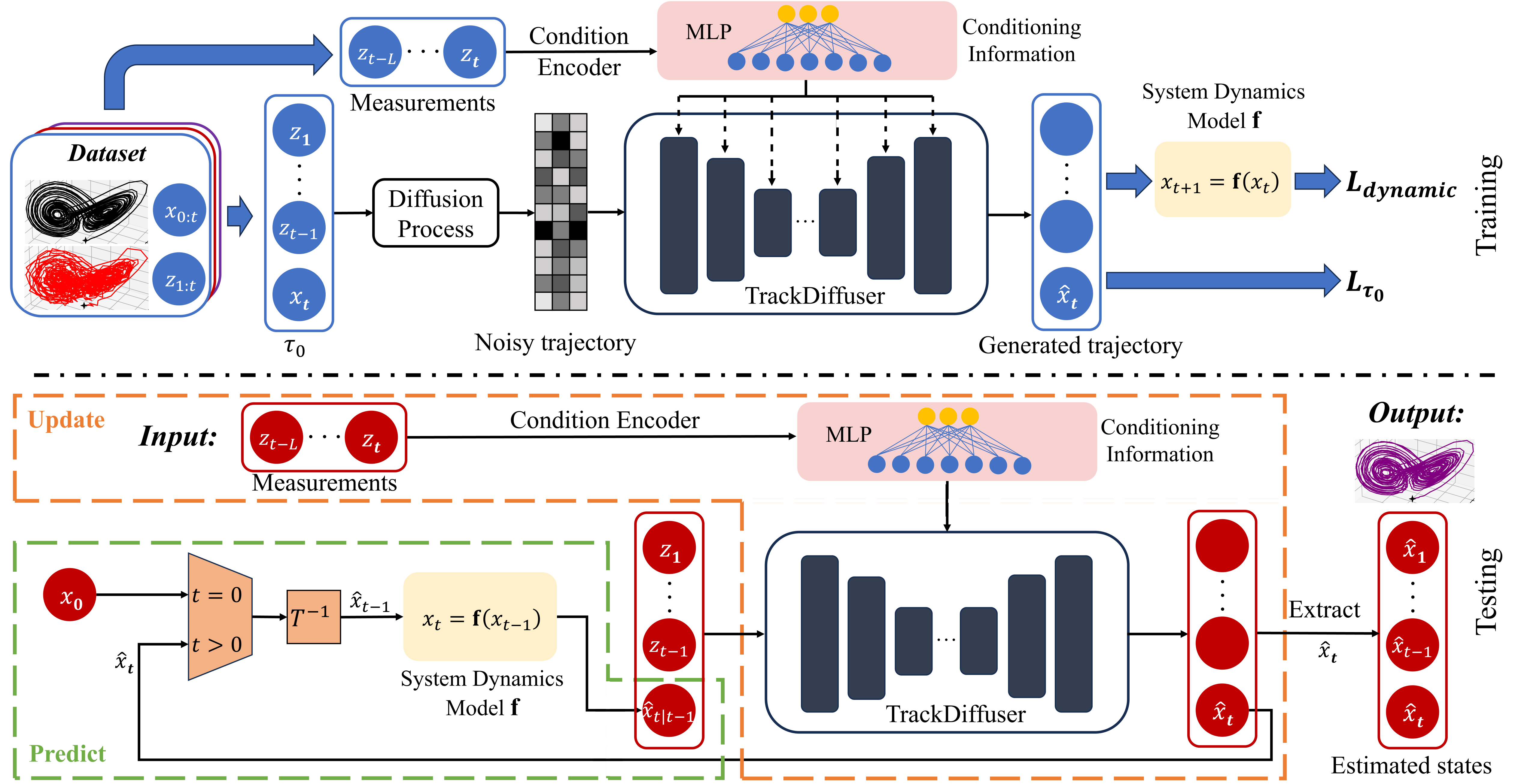}
    \caption{The framework of TrackDiffuser. 
    Bayesian filtering is modeled as a conditional generative modeling problem to approximate the posterior probability density distribution. 
    In training, historical measurements and true states are combined to form a ground-truth trajectory for the diffusion model, and the current measurements sequence is used as a condition for conditional diffusion sampling to obtain state estimates. 
    In inference, we implement the predict step by taking the prior state estimate as the noise mean, and then implement the update step by recursively denoising with measurements conditional guidance. }
    \label{fig1}
\end{figure*}

In this section, we present the design of TrackDiffuser in detail. 
In Figure \ref{fig1}, Bayesian filtering is modeled as a conditional generative modeling problem, to obtain the posterior probability density distribution with a nearly model-free way.
TrackDiffuser combines MB Bayesian filtering framework with a diffusion model to mitigate the effects of inaccurate SSM, while simultaneously circumventing the need for explicit measurement models and noise priors by establishing a direct relationship between measurements and states. 

\subsection{Diffusion over Bayesian Filtering}
%问题定义
In Bayesian filtering, our goal is to recover the state \( x_t \) from the measurements \( z_{1:t} \). 
To mitigate the effects of inaccurate SSM, our approach implicitly learns system dynamics from data. 
The diffusion process is applied across the states \( x_t \) combined with history measurements \( z_{1:t-1} \) as a sequence trajectory, as defined below:

\begin{equation}\label{3.1.1}
{\tau_k}\coloneqq{{(z_1,z_2,\dots,z_{t-1},x_t)}_k},
\end{equation} where $k$ denotes the step in the forward process, and $t$ denotes the time when a state $x_t$ was measured as $z_t$ in true trajectory $\tau_0$. 
$\tau_k$ is treated as a noisy sequence of states from a trajectory of length $t$.
%and represented as a two-dimensional array with one column for a time sequence. 
% a two-dimensional array with one column for a time sequence???两维，一列？
Subsequently, we use the measurements \( z_{1:t} \) as conditions to guide the model in the denoising process. 
Hence, we formulate Bayesian filtering as the standard problem of conditional generative modeling:
\begin{equation}\label{3}
\mathop{\max}_{\theta} \mathbb{E}_{\tau_0 \sim \mathcal{D}}[ \log p_\theta \left( \tau_0 \mid z_{1:t} \right)].
\end{equation}
%main idea的表述感觉有点奇怪，描述measurements作为条件时候
The key idea of TrackDiffuser is to recursively learn the distribution $p_\theta ({\tau_0}\mid{z_{1:t}})$, from true trajectories $\tau_0$ as well as observed measurements, through a diffusion model. Then, we can sample from the learned distribution to obtain posterior state estimate.

\subsection{Filtering with Classifier-Free  Guidance}

Given a diffusion model trained by different trajectories, we introduce how we utilize it for filtering. 
Considering the aim is to learn the distribution $p_\theta({\tau_0}\mid{z_{1:t}})$, it is necessary to additionally condition the diffusion process on measurements $z_{1:t}$. 
One approach is to train a classifier\footnote{We keep the term 'classifier' to align the domain of diffusion model, and the classifier here acturally denotes the guiding predictor.}
 $p_\phi ({z_{1:t}}\mid{\tau_k})$ to predict $z_{1:t}$ from noisy trajectories $\tau_k$, as $ p_\theta(\tau_k\mid z_{1:t}) \propto p_\theta(\tau_k)p_\phi ({z_{1:t}}\mid{\tau_k}) $.
This involves training a neural network to estimate the likelihood distribution, which requires a separate and complex inverse procedure.

One approach to avoid a pre-trained classifier is directly training a conditional diffusion model conditioned with the measurements $z_{1:t}$. 
Formally, to implement classifier-free guidance, the diffusion model is trained jointly under conditional and unconditional model by randomly discarding the conditional label $z_{1:t}$ as follows:
% QQQ
\begin{equation}\label{3.2.2}
\hat{\tau_0}\coloneqq\tau_{0_{\theta}} (\tau_k,\phi,k)+\omega(\tau_{0_{\theta}} (\tau_k,z_{1:t},k)-\tau_{0_{\theta}} (\tau_k,\phi,k)),
\end{equation} where $\omega$ denotes the guidance weight applied to $(\tau_{0_{\theta}} (\tau_k,z_{1:t},k)-\tau_{0_{\theta}} (\tau_k,\phi,k))$. 
With these ingredients, sampling from the TrackDiffuser becomes similar to filtering over measurements. 

\subsection{Conditioning Beyond Measurements}

% \begin{figure}
%     \centering
%     \includegraphics[width=0.8\linewidth]{images/fig2.png}
%     \caption{Caption}
%     \label{fig:enter-label}
% \end{figure}
To circumvent the need for explicit noise priors, we establish a direct relationship between states and measurements.
Specifically, we condition the noise model on the measurements, represented as \( \tau_{0_{\theta}} (\tau_k, z_{1:t}, k) \).
Considering the sensor misalignment situation, we design the conditioning information module to circumvent the need for explicit measurement model.
To preserve the interpretability advantages of traditional model-based filtering methods, We design the system dynamics constraint module.
Based on these two modules, an implicit predict-and-update mechanism is presented.

\paragraph{Conditioning Information}
In practice, sensor misalignment can lead to observation shifts, which is equivalent to the measurement model rotation, potentially causing filter divergence. Therefore, we design the original measurements as conditioning information to guide the denoising process. 
Besides, as \( t \) increases, the condition \( z_{1:t} \) becomes progressively longer. Due to the Markov property, conditioning on \( z_{1:t} \) can introduce redundant information. Therefore, we limit the conditioning to a fixed length \( L \) so \( cond \coloneqq z_{t-L+1:t} \).

\paragraph{System Dynamics Constraint}
Purely data-driven approach \cite{10320373, YAN2024102580} abandons valuable dynamic model structures, struggling with long-term dependencies and system variations and lacking interpretability.
We establish a connection between the system dynamics and the noise sampled during the denoising process. 
Specifically, based on Equation (\ref{2.1.1}), if the process noise follows a Gaussian distribution with a zero mean, we can express \(p(\mathbf{x}_t \mid \mathbf{x}_{t-1}) \sim \mathcal{N}(\mathbf{f}(\mathbf{x}_{t-1}), Q_t)\). Even under non-Gaussian noise, the current state \(\mathbf{x}_t\) follows a distribution with a mean of \(\mathbf{f}(\mathbf{x}_{t-1})\). Therefore, in our study, we can bypass the assumption in diffusion models that the final noisy trajectory follows a Gaussian distribution with zero mean. Instead, we now assume that the noisy trajectory follows a Gaussian distribution with a mean of \(\mathbf{f}(\mathbf{x}_{t-1})\).
In addition, we introduce terms related to the system dynamics constraint in the design of the loss function.

\paragraph{Predict and Update}
We have learned the data distributions \( p(\tau_0 \mid cond) \), i.e., \( p({x}_t \mid {z}_{1:t}) \). 
Filtering then becomes the process of sampling a trajectory conditioned on \( cond \). 
First, we observe the measurement \( z_t \) and extract the conditional input \( cond \) based on the conditional length \( L \). Next, we compute the prior state estimate as the mean of the noise sampled during the denoising process. Finally, we sample a trajectory to obtain the posterior state estimate, which constitutes the update step. We observed that low-quality data could potentially be generated during sampling, and we adopt the low-temperature sampling method \cite{ajay2023is} to address this issue. This procedure iteratively follows the Bayesian filtering framework, as outlined in Algorithm 1.

\begin{algorithm}[tb]
    \caption{Bayesian filtering with TrackDiffuser}
    \label{algorithm 1}
\begin{algorithmic}
    \STATE \textbf{Input:} noise model \( \tau_{0_\theta} \), guidance scale \( \omega \), history length \( T \), initial state \( x_0 \), condition length \( L \)
    \STATE \textbf{Output:} Estimated states queue \( h \)
    \STATE Initialize \( h \leftarrow \) Queue(length = \( T \)), \( Z \leftarrow \) Queue(length = \( T \)), \( t \leftarrow 1 \)
    \WHILE{not done}
    \STATE Observe measurement \( z_t \); \( Z \).insert(\( z_t \)); Initialize \( \tau_K \sim N(0, \alpha I) \)
    \FOR{\( k = K, \dots, 1 \)}
    \STATE \( \tau_k[:\text{length}(h)-1] \leftarrow Z[:\text{length}(h)-1] \); 
    \STATE \( \tau_k[h] \leftarrow \tau_k[h] + \mathbf{f}(x_{t-1}) \) \texttt{ \# predict step}
    \STATE \( cond \leftarrow Z[-L:-1] \)
    \STATE \( \hat{\tau_0} \leftarrow (1-\omega)\tau_{0_\theta} (\tau_k, \phi, k) + \omega (\tau_{0_\theta} (\tau_k, cond, k) \)
    \STATE \( (\mu_{k-1}, \sum_{k-1}) \leftarrow \text{Denoise}(\tau_k, \hat{\tau_0}) \)
    \STATE \( \tau_{k-1} \sim N(\mu_{k-1}, \alpha \sum_{k-1}) \) \texttt{ \# update step}
    \ENDFOR
    \STATE Extract \( x_t \) from \( \tau_0 \)
    \STATE \( h \).insert(\( x_t \)); \( t \leftarrow t + 1 \)
    \ENDWHILE
\end{algorithmic}
\end{algorithm}

\subsection{Training}

For training the conditional generative model, we collect a set of \( \left\{(z_{1:T_{i}}^i,x_{0:T_{i}}^i)\right\}_{i=1}^N \) of trajectories from a determined SSM. These trajectories are then processed into samples as dataset \( \mathcal{D}=(\tau_0 = (z_1, z_2, \dots, z_{t-1}, x_t), cond = z_{t-L+1:t}) \), with each trajectory having a varying length.
Like \cite{jun2023shapegeneratingconditional3d, venkatraman2024reasoning}, we found that initial $\tau_0$ predict mode yields better results than the noise predict mode in our experiments. We train the reverse diffusion process $p_{\theta}$, incorporating system dynamics constraint, parameterized through $\tau_{0_{\theta}}$ as follow:

% \begin{equation}\label{3.4.1}
%     \mathcal{L}(\theta,\phi) : = E _ { k,\tau_0 \sim \mathcal{D} , \beta \sim \text{Bern}(p)} \left[ \mid \mid \epsilon - \epsilon _ { \theta } ( \tau_k , (1-\beta)cond+\beta\phi,k ) \mid \mid ^ { 2 } \right]
% \end{equation}
\begin{equation}\label{3.4.1}
    \begin{aligned}
    \mathcal{L}_{total} &= \mathcal{L}_{\tau_0} + \mathcal{L}_{dynamic},\\
     \mathcal{L}_{\tau_0}(\theta,\phi) &= E \left[ \|\tau_0 - \tau_{0_{\theta}}(\tau_k, (1-\beta)cond+\beta\phi, k)\|^2 \right],\\
    \mathcal{L}_{dynamic}(\theta) &= E \left[ \|\mathbf{f}(\tau_0)-\mathbf{f}(\tau_{0_{\theta}})\|^2 \right],
    %+E \left[ \|f(x_{0:T_{i}}^i) - f(x_{0:T_{i}}^i (\theta))\|^2 \right],    
    \end{aligned}
\end{equation} where the loss \(\mathcal{L}_{dynamic}(\theta)\) evaluates the consistency of the predicted trajectory with the true system dynamics. Specifically, it additionally introduces information about the non-linear motion of the SSM. The expectation is taken over timestep samples $k$, trajectory samples $\tau_0 \sim \mathcal{D}$, and Bernoulli random variable $\beta \sim \text{Bern}(p)$. This combined loss ensures that the predicted trajectory not only aligns with the ground truth but also adheres to the correct dynamic evolution over time, preserving the underlying system dynamics.

The overall framework of TrackDiffuser is shown in Figure \ref{fig1}. During training, ground truth trajectories are formed from historical measurements and states truth as input to the diffusion model, while current measurement sequences serve as conditions for conditional diffusion sampling to estimate the state. In more detail, we parameterize $\tau_{0_\theta}$ with a temporal U-Net architecture, a neural network consisting of repeated convolutional residual blocks \cite{ajay2023is}. This architecture processes the trajectory $\tau_k$ into a two-dimensional vector, with the width corresponding to the state dimension and the length reflecting the time horizon. The conditioning information, $z_{t-L+1:t}$, is encoded as a learned embedding, creating a connection between the observed measurements and the true state. A key component, the Residual Temporal Block, captures temporal dependencies by adapting to the variable length of the observed trajectory. Through its convolutional structure, the block processes both the current state $\tau_0$ and the measurement sequence $z_{t-L+1:t}$, allowing the model to effectively integrate information across multiple time steps.

The architecture thus provides a robust framework for fusing both spatial(system dynamics constraint) and temporal information(conditioning information), capturing complex dynamics between the observed measurements and the underlying states. 

\section{Experiments}
In this section, we evaluate the performance of the TrackDiffuser on a non-linear state-space model under multiple experimental setups:
1) We investigate the performance under Gaussian noise, testing both linear and non-linear measurement scenarios.
2) We examine the effects of non-Gaussian noise, specifically Gaussian mixture noise, across linear and non-linear measurement settings.
3) In real-world applications, the dynamics model is often not fully known and is usually based on prior assumptions, which may lead to mismatch conditions. We evaluate performances under SSM mismatch and training-testing mismatch scenarios.
4) Finally, we evaluate the effectiveness of TrackDiffuser in real-world applications using the Michigan NCLT data set \cite{10.1177/0278364915614638}.
To ensure a comprehensive assessment, we compared TrackDiffuser with traditional model-based methods, such as EKF, UKF, and PF, as well as the hybrid method, KalmanNet \cite{9733186}.

\subsection{Experimental Setting}

Lorenz attractor is a three-dimensional chaotic solution derived from the Lorenz system of ordinary differential equations in continuous time \cite{Lorenz2004}. It effectively captures the dynamics of complex systems exhibiting high sensitivity and chaotic behavior, providing a challenging yet valuable benchmark for testing filtering algorithms in non-linear dynamics. The solution of the continuous Lorenz attractor over a short time interval $\Delta$ is given by: 
\begin{equation}\label{6.0.1}
    {x}_{t} = \text{exp}(\mathbf{A}( {x}_{t-1}) \cdot \Delta) \cdot {x}_{t-1},\mathbf{A}( {x}_{t-1})=\begin{bmatrix}
-10 & 10 & 0 \\
28 & -1 & -{x}_{t-1, 1} \\
0 & {x}_{t-1, 1} & -\frac{8}{3}
\end{bmatrix}.
\end{equation}
As a result, the SSM is in a discretized time-series form as: 
\begin{equation}\label{4.0.1}
    {x}_t = \mathbf{F}({x}_{t-1}){x}_{t-1} + \mathbf{w}_t,
\end{equation} where the system dynamics is a non-linear function of state $x_{t-1}$, the process noise \( \mathbf{w}_t \) here is modeled as Gaussian and non-Gaussian noise, respectively.
\begin{equation}\label{4.0.2}
    \mathbf{F}({x}_{t-1}) = \exp(
    %\left( 
    \begin{bmatrix}
    -10 & 10 & 0 \\
    28 & -1 & -{x}_{t-1, 1} \\
    0 & {x}_{t-1, 1} & -\frac{8}{3}
    \end{bmatrix}
    \cdot
    \Delta
    %\right)
    ).
\end{equation}

Considering the general practice in the field of non-linear filtering, unlike KalmanNet \cite{9733186}, we directly use the state-evolution function, i.e., Equation (\ref{4.0.2}) with the step-size $\Delta$= 0.02 seconds to generate simulation data.

\paragraph{Finite-Taylor series approximation} \label{Appendix_A}
In mismatched conditions, we take finite-Taylor series approximation with \( J \) coefficients of (\ref{6.0.1}), which leads to (\ref{4.0.2}) being approximated as:

\begin{equation}\label{6.0.2}
    \mathbf{F}({x}_{t-1}) = \text{exp}(\mathbf{A}( {x}_{t-1}) \cdot \Delta) \approx \mathbf{I}+ \sum_{j=1}^{J} \frac{(\mathbf{A}( {x}_{t-1}) \cdot \Delta)^j}{j!}.
\end{equation}

\paragraph{Gaussian noise} 
The noise is modeled as a normal distribution, i.e., process noise \( \mathbf{w}_t \sim \mathcal{N}(0, \mathbf{Q}) \) and measurement noise \( \mathbf{v}_t \sim \mathcal{N}(0, \mathbf{R}) \).

\paragraph{Non-Gaussian noise} 
Considering common approaches to modeling non-Gaussian noise, we model the noise process as a Gaussian mixture (GM) distribution, i.e., process noise \( \mathbf{w}_t \sim \alpha \mathcal{N}(0, \mathbf{Q_1}) + (1-\alpha) \mathcal{N}(0, \mathbf{Q_2}) \) and measurement noise \( \mathbf{v}_t \sim \alpha \mathcal{N}(0, \mathbf{R_1}) + (1-\alpha) \mathcal{N}(0, \mathbf{R_2}) \).

The same numerical study setup as that in KalmanNet \cite{9733186} is considered in the simulation, where the SSM is generated using diagonal noise covariance matrices: 
\begin{equation}\label{4.0.3}
    \mathbf{Q} = \mathrm{q}  ^ { 2 } \cdot \mathbf{I} , \mathbf{R} = \mathrm{r} ^ { 2 } \cdot \mathbf{I} , \nu \triangleq \frac{ \mathrm{q} ^ { 2 } } { \mathrm{r} ^ { 2 } }, 
\end{equation} where \(\nu\) represents the ratio of the process noise variance to the measurement noise variance. In this paper, the metric used to evaluate performance is the MSE on a [dB] scale. In the table, we present the MSE (in [dB]) as a function of the inverse measurement noise level, i.e., \( 1 / \mathrm{r} ^ { 2 }\), which is also on a [dB] scale.

In all experiments TrackDiffuser is trained using the Adam optimizer. 
We implemented EKF, UKF and PF using PyTorch. 
We used the implementation of KalmanNet from \cite{9733186}. 
If not specified, the Jacobian matrix for both EKF and KalmanNet is computed using a 5th-order Taylor expansion. 
Additionally, the linear measurement refers to the identity transformation, while the non-linear measurement involves the transformation from Cartesian to spherical coordinates.

\begin{figure}[htb!]
\centering

\subfigure[\centering EKF]{
    \includegraphics[width=0.26\linewidth]{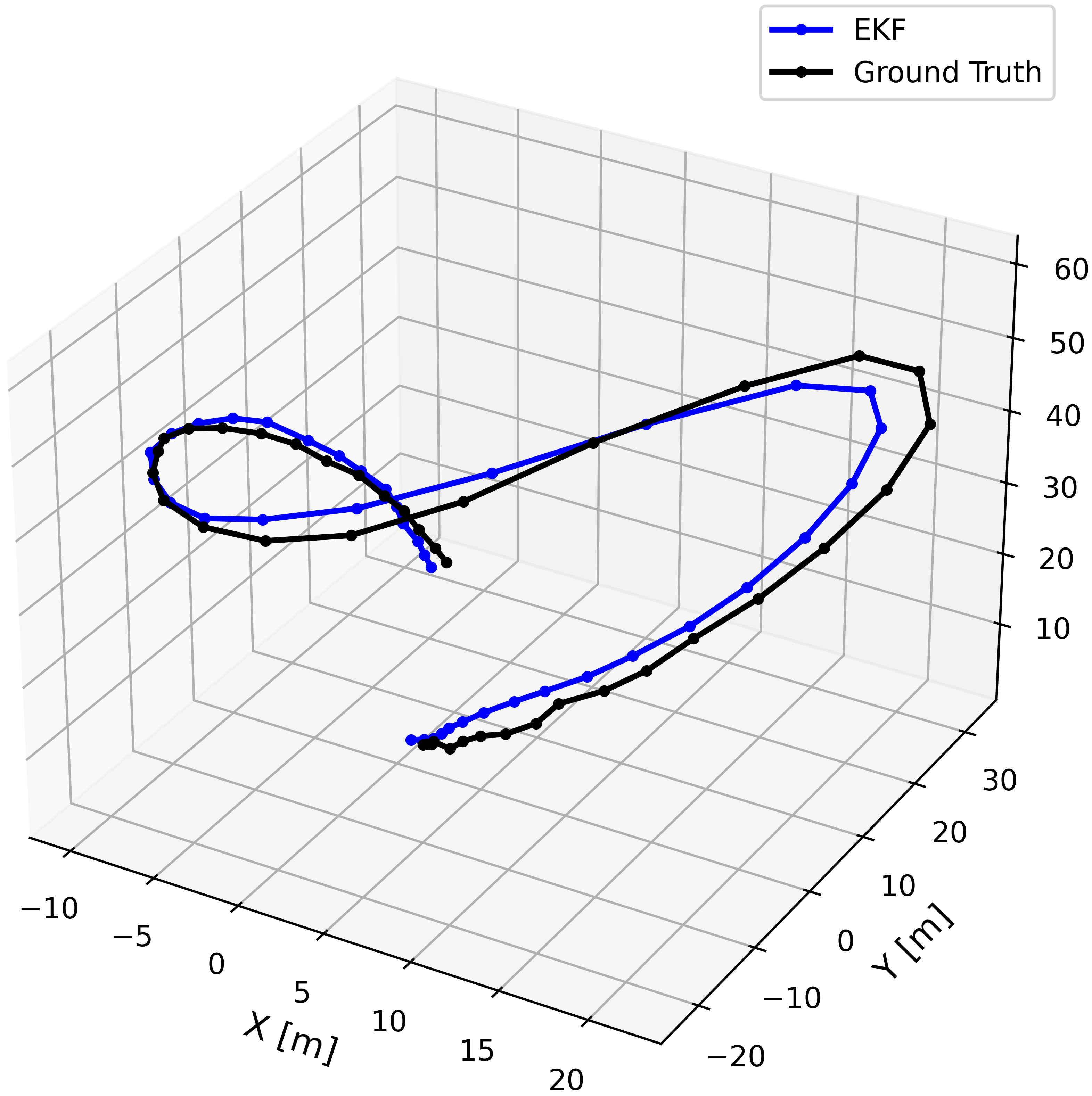}
    \label{fig:EKF}
}
\subfigure[\centering UKF]{
    \includegraphics[width=0.26\linewidth]{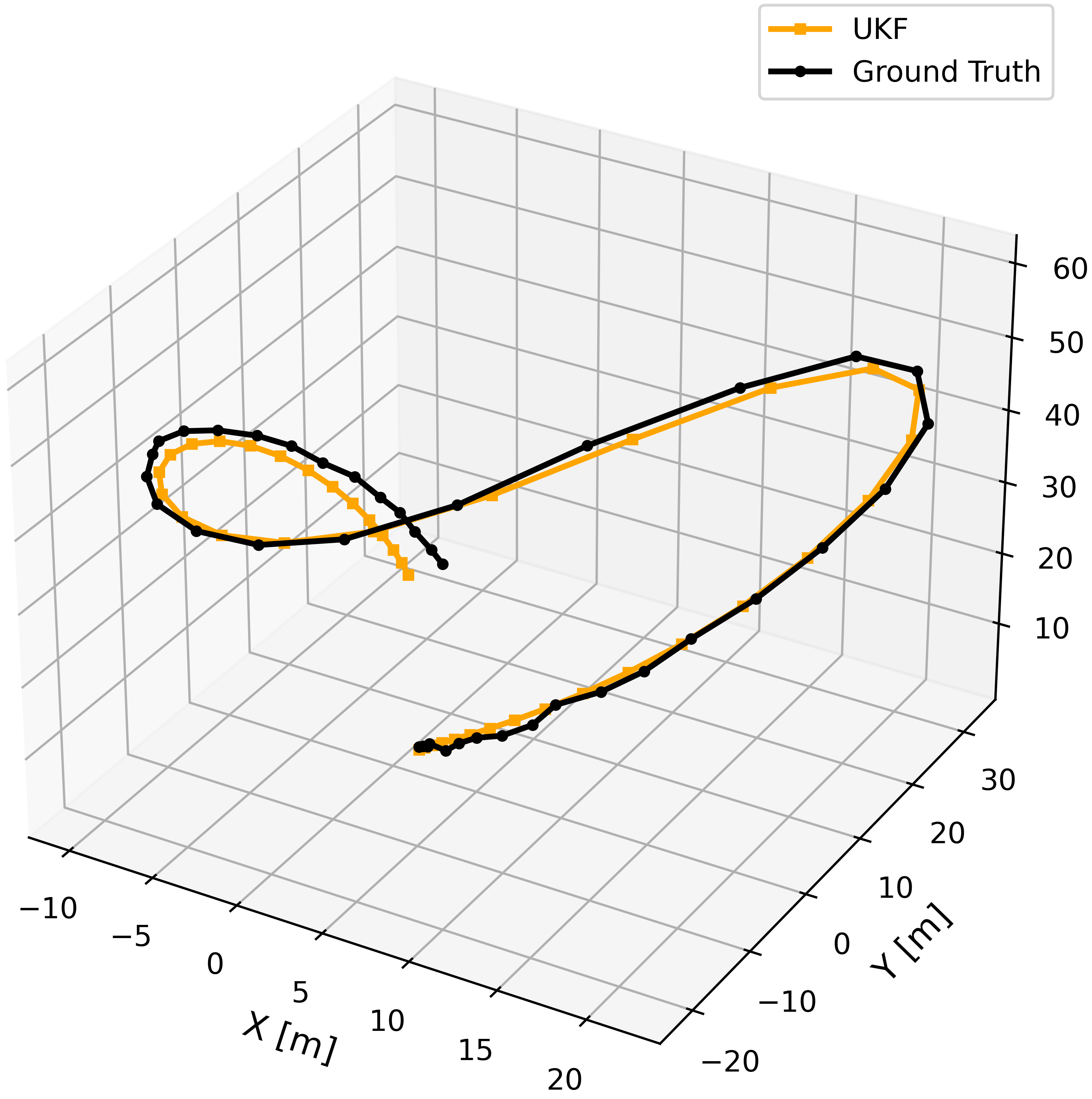}
    \label{fig:UKF}
}
\subfigure[\centering PF]{
    \includegraphics[width=0.26\linewidth]{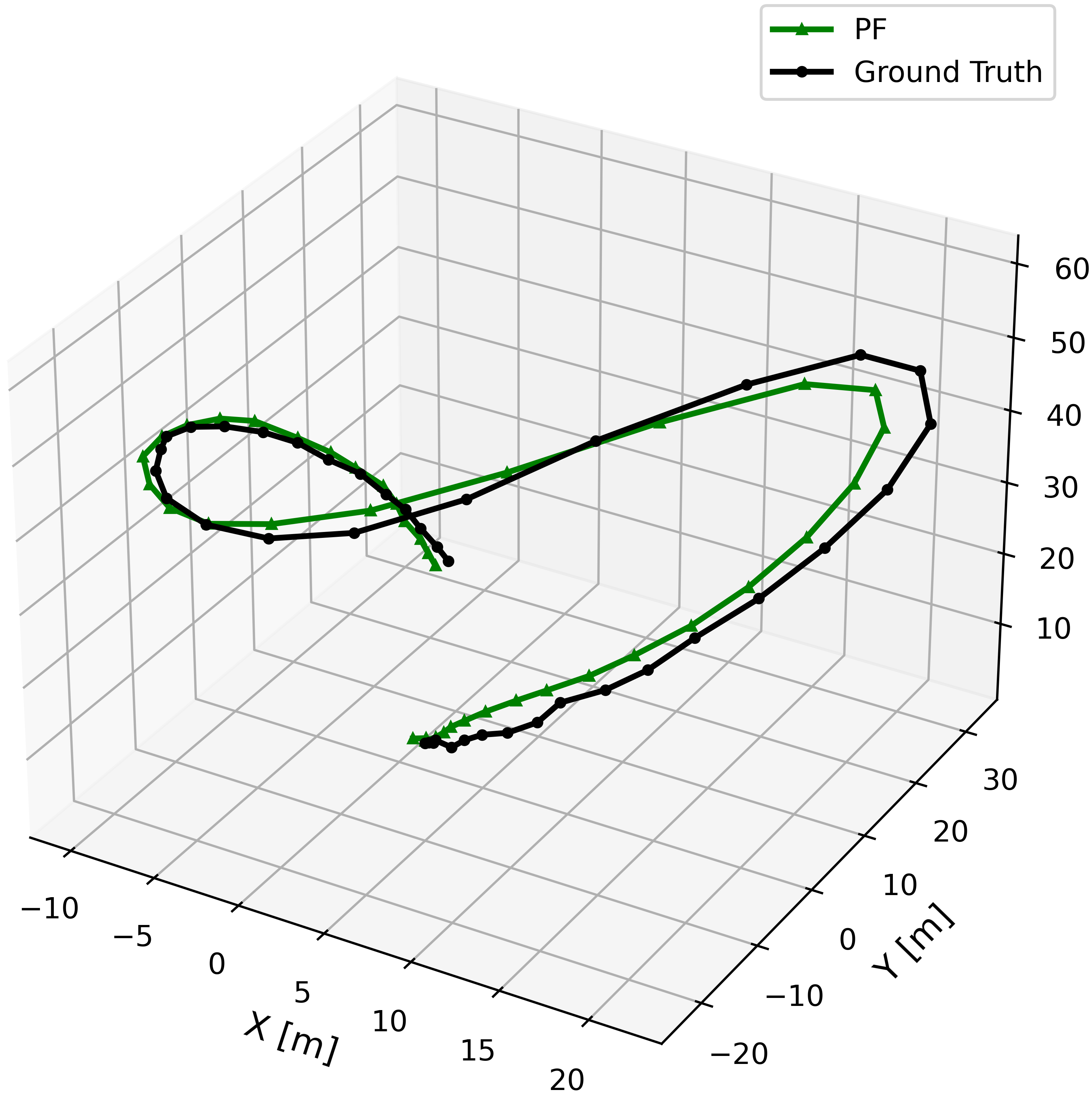}
    \label{fig:PF}
}
\subfigure[\centering KalmanNet]{
    \includegraphics[width=0.26\linewidth]{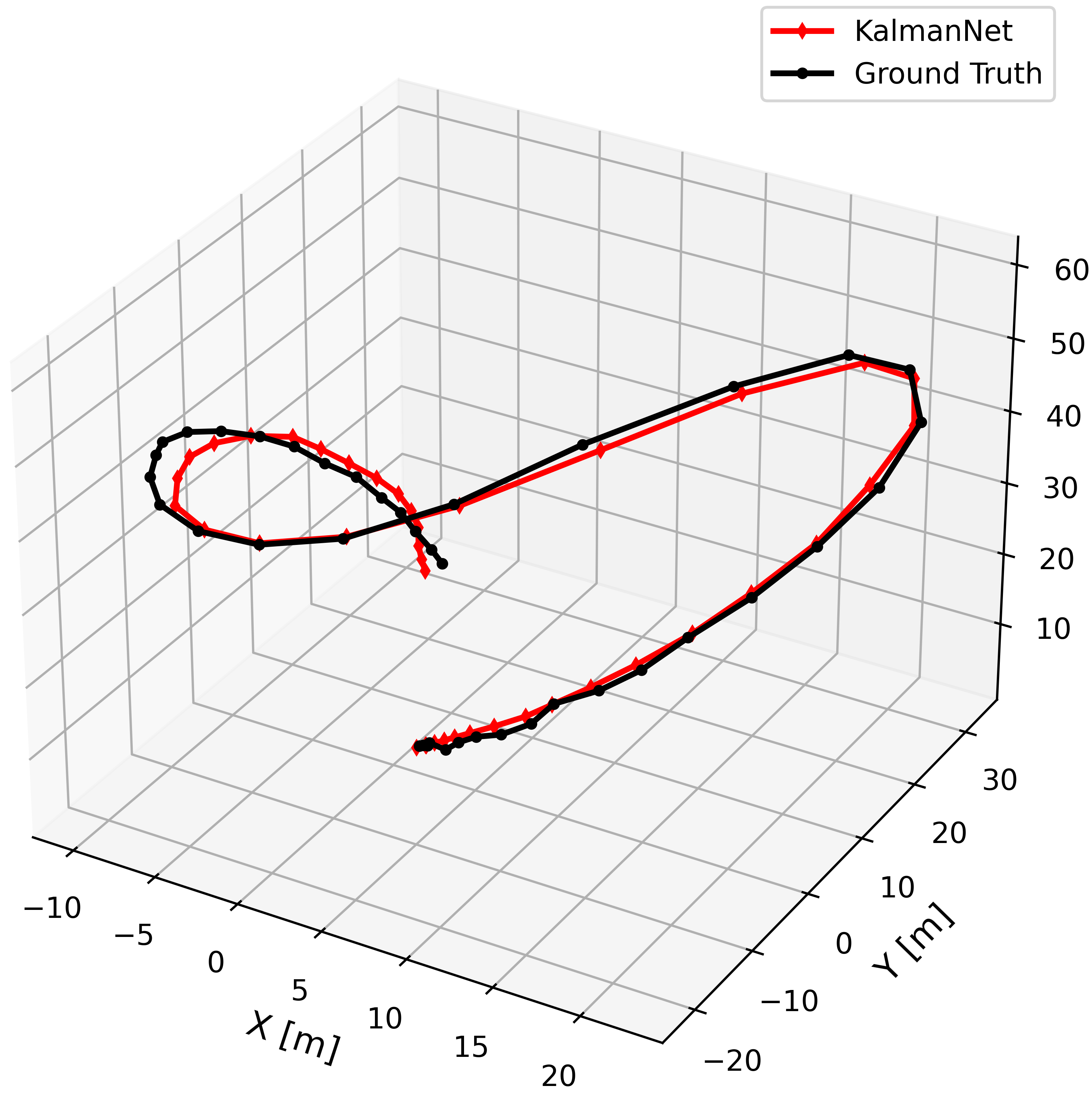}
    \label{fig:KalmanNet}
}
\subfigure[\centering TrackDiffuser]{
    \includegraphics[width=0.26\linewidth]{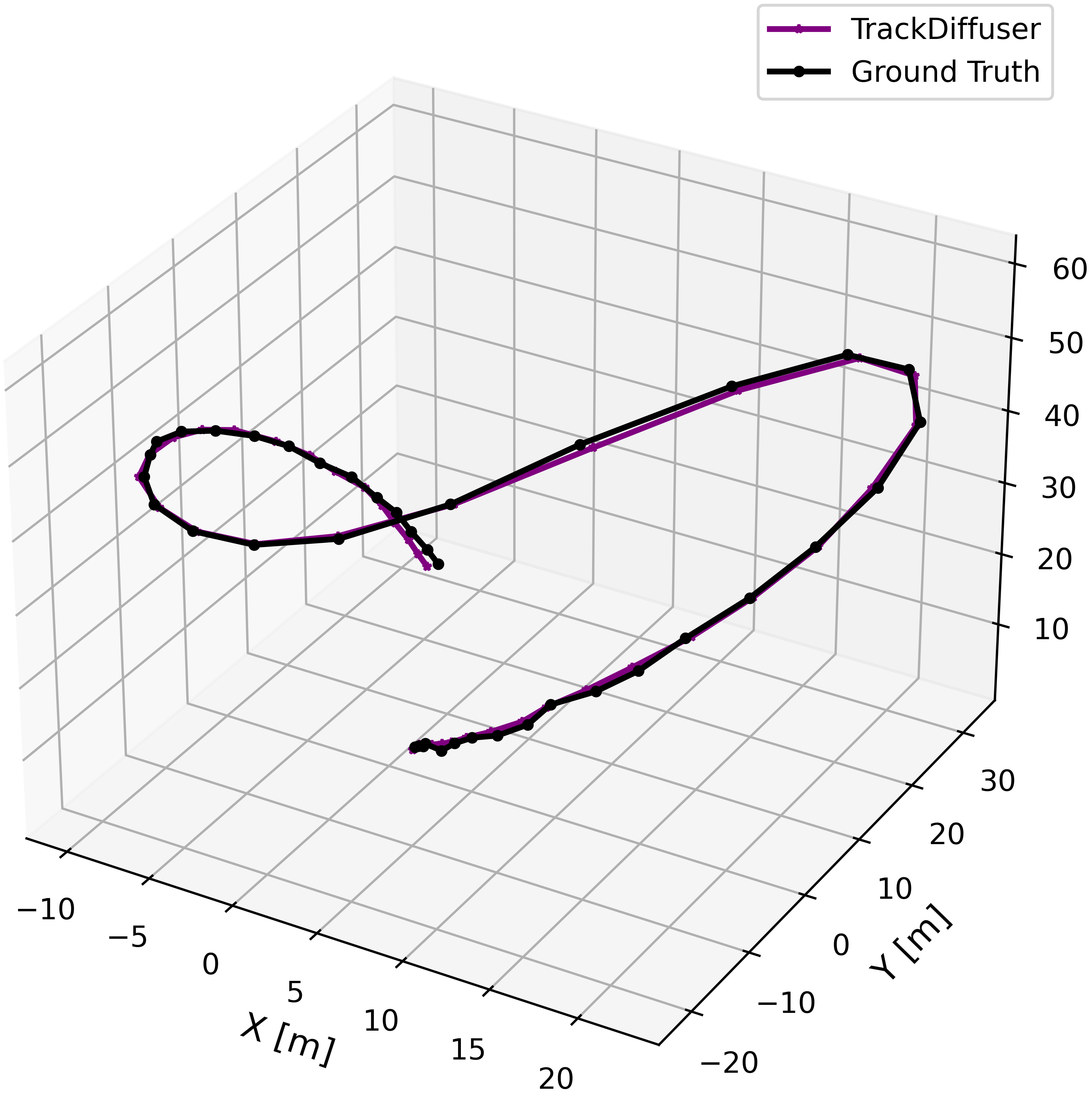}
    \label{fig:TrackDiffuser}    
\label{fig2}
}
\caption{Trajectory visualization of filtering Lorenz attractor SSM with non-linear measurements, $\mathrm{q}^2$ = $\mathrm{r}^2$ = -10 [dB], $T = 40$. TrackDiffuser performs the best, especially in manoeuvring corners.}

\end{figure}

\subsection{Gaussian State Space Model}

Our first experiment is conducted under Gaussian noise, comparing the performance with other filters under linear and non-linear measurements settings.

%%% QQQ
\paragraph{Linear measurements with Gaussian noise}
we set \( \nu = -20 \, \text{[dB]} \) and \( T = 100 \). As shown in Table \ref{table1}, TrackDiffuser demonstrated excellent MSE performance, outperformed EKF, UKF, PF, and KalmanNet.

\paragraph{Non-linear measurements with Gaussian noise}
Next, we set \( \nu = 0 \, \text{[dB]} \) and \( T = 20 \). As shown in Table \ref{table2}, we observe that hybrid methods both KalmanNet and TrackDiffuser can mitigate the interference caused by non-linear measurements, outperforming MB methods, with TrackDiffuser achieving the best MSE.

% \begin{figure*}
%     \centering
%     \includegraphics[width=0.8\linewidth]{images/fig3.trajectoriesR1J5.png}
%     \caption{Filtering for Lorenz attractor SSM using TrackDiffuser, $T = 100$. The true state trajectory at $\mathrm{q}^2$ = -20 [dB] is shown on the left, noisy measurement trajectory at $\mathrm{r}^2$ = 0 [dB] in the middle, the estimated trajectory on the right.}
%     \label{fig3}
% \end{figure*}

\begin{table}[htb!]
\centering
\caption{MSE - Linear measurements with Gaussian noise}
\label{table1}
\begin{tabular}{|c|c|c|c|c|c|c|}
\hline
$1/\mathrm{r}^2 \, \textbf{[dB]}$ & \textbf{-10} $\downarrow$ & \textbf{0} $\downarrow$ & \textbf{10} $\downarrow$ & \textbf{20} $\downarrow$  & \textbf{30} $\downarrow$ \\
\hline
EKF  & 2.69 & -6.19 & -16.49 & -25.18 & -34.81 \\
\hline
UKF  & 9.05 & 2.58 & -3.52 & -16.24 & -23.64 \\
\hline
PF  & 3.76 & -4.76 & -14.68 & -22.93 & -29.14  \\
\hline
KalmanNet  & 1.16  & -7.25 & -17.86 & -25.75 & -35.43 \\
\hline
TrackDiffuser & \textbf{1.14} & \textbf{-7.83} & \textbf{-18.23} & \textbf{-26.84} & \textbf{-35.58} \\
\hline
\end{tabular}%
\end{table}

\begin{table}[htb!]
\centering
\caption{MSE - Non-linear measurements with Gaussian noise}
\label{table2}
\begin{tabular}{|c|c|c|c|c|c|}
\hline
$1/\mathrm{r}^2 \, \textbf{[dB]}$ & \textbf{-10} $\downarrow$ & \textbf{0} $\downarrow$ & \textbf{10} $\downarrow$ & \textbf{20} $\downarrow$  & \textbf{30} $\downarrow$ \\
\hline
EKF  & 24.25 & 18.46 & 9.27 & -4.14 & -15.17 \\          
\hline
UKF  & 27.90 & 23.59 & 14.89 & 4.88 & -6.46 \\          
\hline
PF  & 23.66  & 15.31 & 10.24  & 0.49 & -9.17 \\
\hline
KalmanNet & 23.51 & 8.06 & -4.37 & -12.94 & -21.65 \\    
\hline
TrackDiffuser & \textbf{20.07} & \textbf{7.12} & \textbf{-5.08} & \textbf{-13.67} & \textbf{-23.93} \\  
\hline
\end{tabular}%
\end{table}

\subsection{Non-Gaussian State Space Model}

Then, we conducted experiments under non-Gaussian noise, where the noise process is modeled as a Gaussian mixture (GM) distribution with \( \alpha = 0.8 \).

\paragraph{Linear measurements with GM noise}
we set \( \nu = -20 \, \text{[dB]} \) and \( T = 100 \). As shown in Table \ref{table3}, TrackDiffuser demonstrated excellent MSE performance, and outperformed PF and other filters.

%需要进一步分析结果，说明我们方法鲁棒性
\paragraph{Non-linear measurements with GM noise}
Next, we set \( \nu = 0 \, \text{[dB]} \) and \( T = 20 \). In line with the non-linear measurements with Gaussian noise case, TrackDiffuser demonstrated the best capability in handling non-linear measurements. Furthermore, comparing Table \ref{table4} with Table \ref{table2}, it can be found that with the variance of the GM noise increases, the performance of all filtering methods deteriorates significantly. 

\begin{table}[htb!]
\centering
\caption{MSE - Linear measurements with Non-Gaussian noise}
\label{table3}
\begin{tabular}{|c|c|c|c|c|c|}
\hline
$1/\mathrm{r}^2 \, \textbf{[dB]}$ & \textbf{-10} $\downarrow$ & \textbf{0} $\downarrow$ & \textbf{10} $\downarrow$ & \textbf{20} $\downarrow$  & \textbf{30} $\downarrow$ \\
\hline
EKF  & 7.89 & 0.07 & -9.52 & -19.32 & -26.77 \\                  
\hline
UKF  & 19.16 & 10.84 & 2.50 & -9.67 & -20.41 \\                  
\hline
PF  & 8.13  & 1.16  & -9.41  & -19.93 & -27.46 \\
\hline
KalmanNet & 7.66  & -0.15 & -10.21 & -18.93 & -26.43 \\     
\hline
TrackDiffuser & \textbf{7.43} & \textbf{-0.66} & \textbf{-10.94} & \textbf{-20.99} & \textbf{-29.20} \\           
\hline
\end{tabular}%
\end{table}

\begin{table}[htb!]
\centering
\caption{MSE - Non-linear measurements with Non-Gaussian noise}
\label{table4}
\begin{tabular}{|c|c|c|c|c|c|}
\hline
$1/\mathrm{r}^2 \, \textbf{[dB]}$ & \textbf{-10} $\downarrow$ & \textbf{0} $\downarrow$ & \textbf{10} $\downarrow$ & \textbf{20} $\downarrow$  & \textbf{30} $\downarrow$ \\
\hline
EKF  & 30.35 & 23.78 & 16.40 & 11.78  & -7.35 \\                
\hline
UKF  & 33.36 & 27.31 & 19.22 & 10.88 & 3.54 \\                
\hline
PF  & 29.03  & 21.47  & 14.92  & 7.36  & -11.95 \\
\hline
KalmanNet & 29.70  & 19.82  & 2.68  & -7.50 & -16.86  \\ 
\hline
TrackDiffuser & \textbf{28.56} & \textbf{15.53} & \textbf{2.03} & \textbf{-8.03} & \textbf{-17.53} \\  
\hline
\end{tabular}%
\end{table}

\subsection{On Mismatched Conditions}
%考虑删除Training Data Amount的研究，删除Mismatched Conditions的四张表格，采用四个子图替代
%H mismatch UKF、PF也使用了h，所以会有偏移影响，但我们方法不使用任何h知识

Given the promising performance of TrackDiffuser, we address two important questions: (a) All experiments mentioned so far have been conducted under the accurate SSM assumption during both the training and testing phases. However, when the underlying system dynamics are not fully understood or cannot be perfectly captured, approximations or simplified models are used, leading to an inaccurate SSM, which results in an SSM mismatch. (b) After training on an accurate system dynamics model $\mathbf{f}(\cdot)$, the system is often adapted in practice to meet changing requirements, such as computational speed. For instance, in KalmanNet and TrackDiffuser, an accurate state-evolution function is used during training, but during testing, a second-order Taylor expansion approximation is applied. In such cases of Training-Testing Mismatch, is the hybrid approach still effective? What happens when these mismatches occur? This question is central to robustness studies.

\begin{figure}[htb!]
\centering

\subfigure[Gaussian System dynamics mismatch]{
    \includegraphics[width=0.4\linewidth]{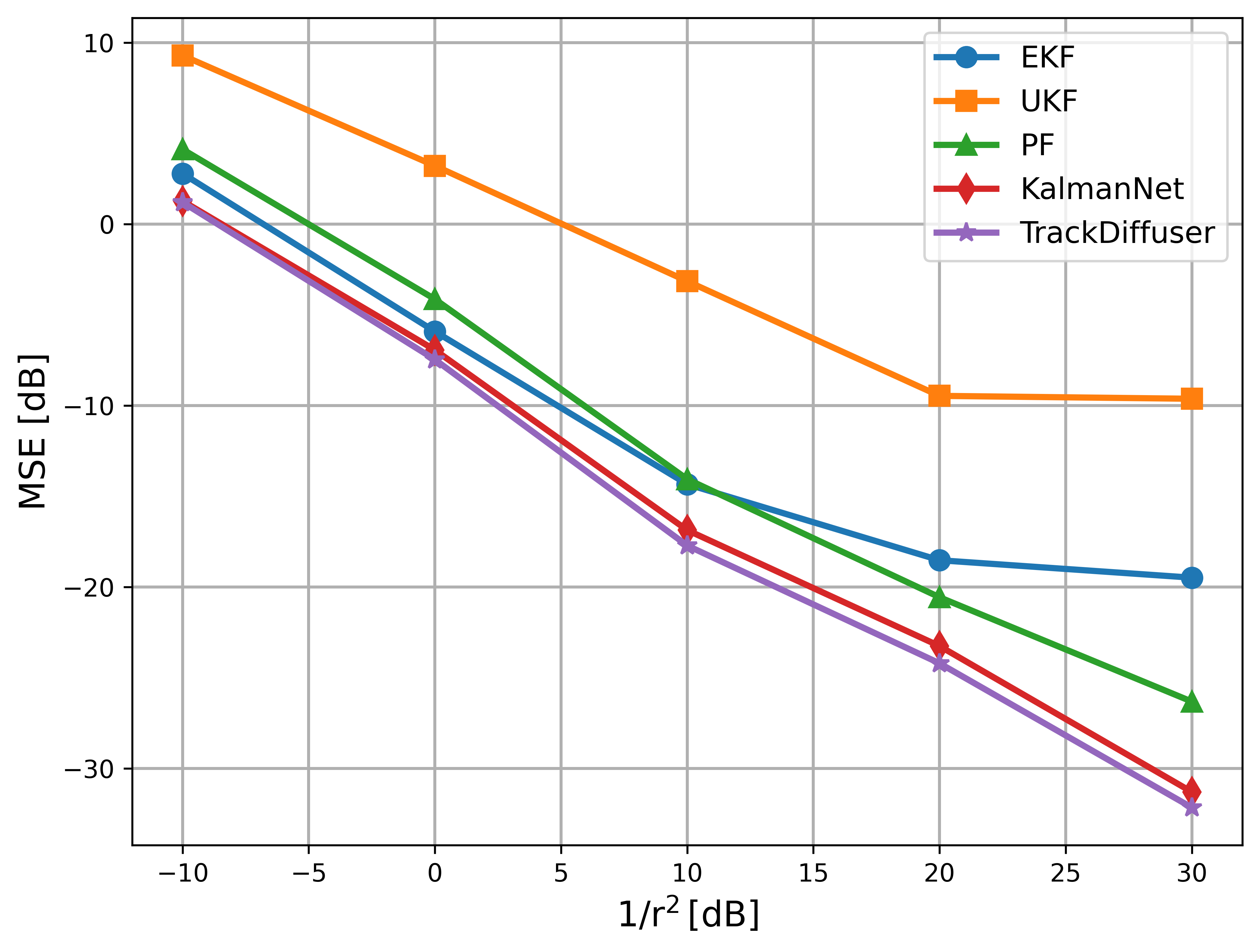}
    \label{mismatchJ2_G}
}
\subfigure[Non-Gaussian System dynamics mismatch]{
    \includegraphics[width=0.4\linewidth]{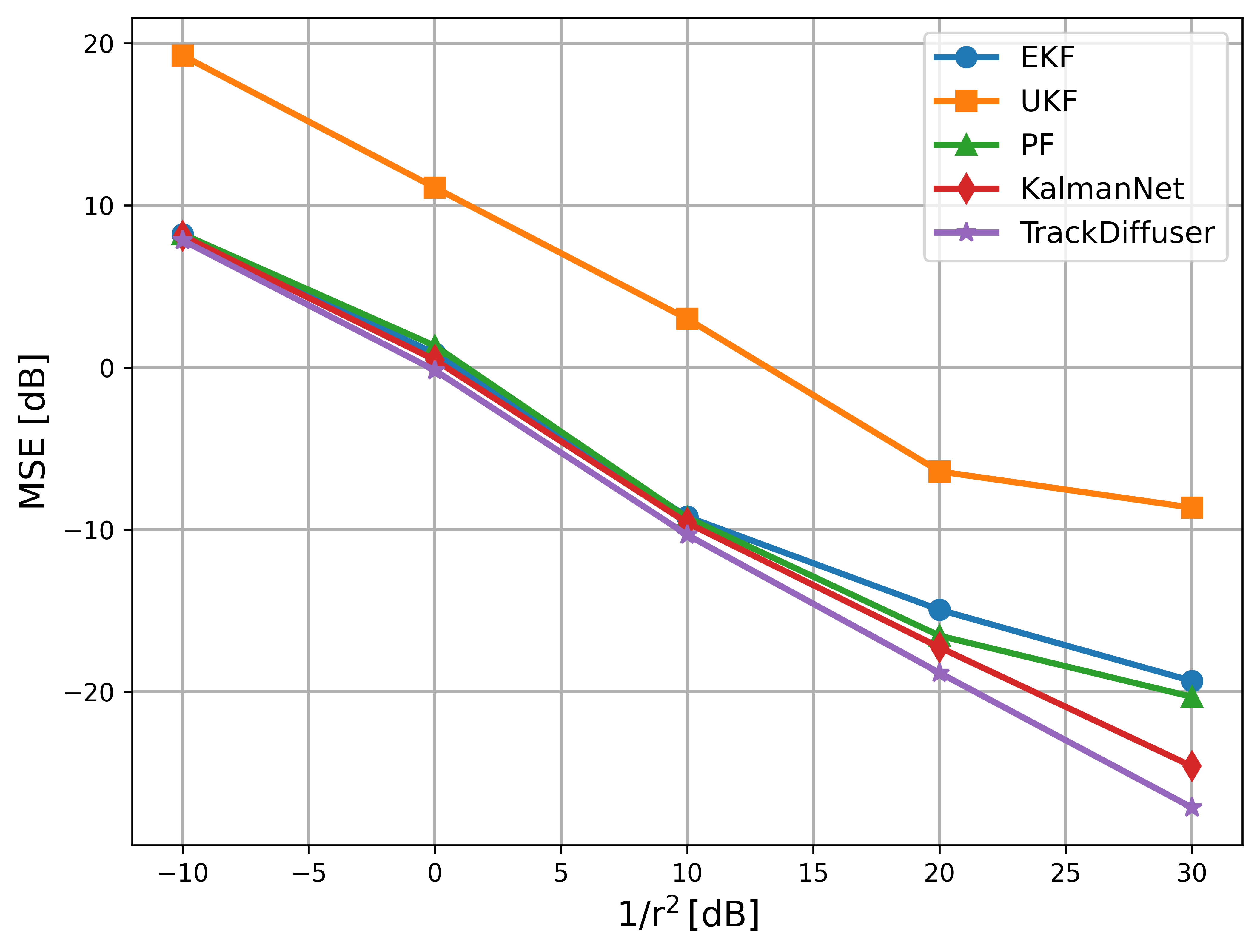}
    \label{mismatchJ2_NG}
}
\subfigure[Gaussian Measurement mismatch]{
    \includegraphics[width=0.4\linewidth]{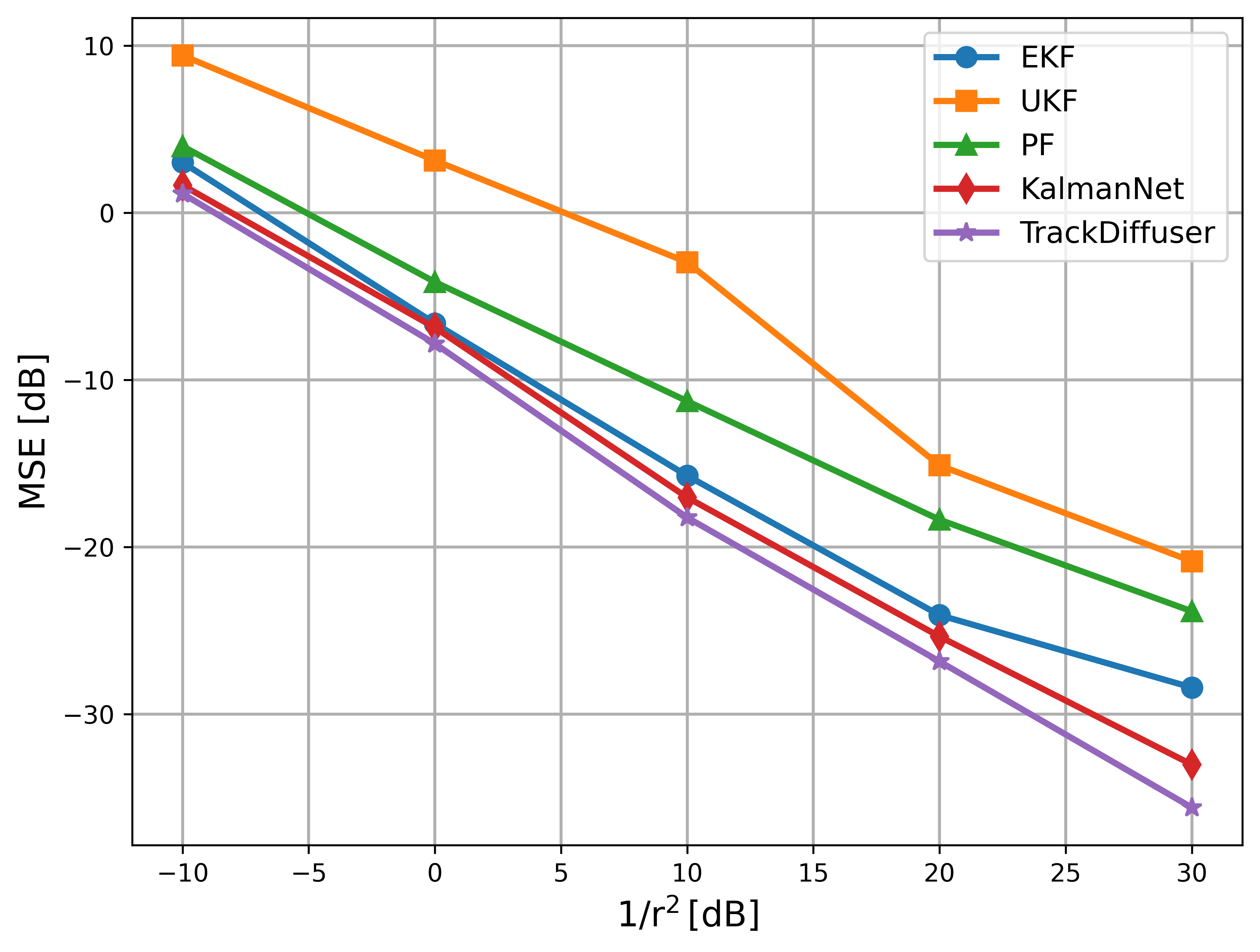}
    \label{mismatchRo_G}
}
\subfigure[Non-Gaussian Measurement mismatch]{
    \includegraphics[width=0.4\linewidth]{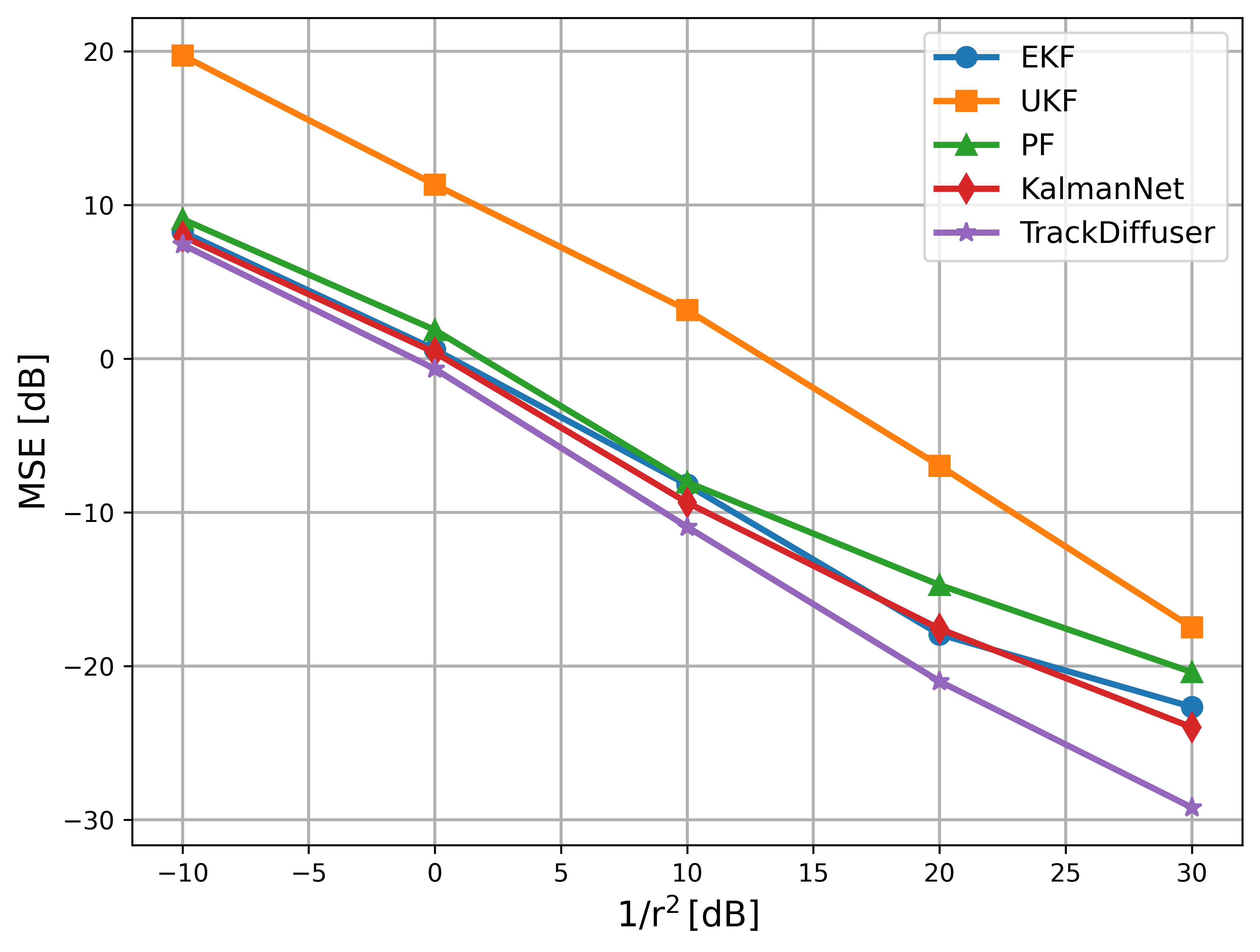}
    \label{mismatchRo_NG}
\label{fig8}
}
\caption{Performance under SSM Mismatch. TrackDiffuser demonstrates superior robustness under system dynamics and measurement rotation mismatches.}

\end{figure}

\subsubsection{Robustness under SSM mismatch}

we proceed to evaluate TrackDiffuser and compare it with other filters under SSM mismatch conditions. 
We study two cases as follows:

%QQQ
\paragraph{System dynamics mismatch}
Using a Taylor series approximation of insufficient order leads to system dynamics mismatch. 
Specifically, we use an approximation with \( J = 2 \) coefficients of Equation (\ref{6.0.2}) to compute the state-evolution function \( \mathbf{f}(\cdot) \) for training and testing, while the data is generated using accurate system dynamics model (\ref{4.0.2}). We again set \( \mathbf{h}(\cdot) \) as the identity mapping, with \( \nu = -20 \, \text{[dB]} \) and \( T = 100 \). The results shown in Figure \ref{mismatchJ2_G} and \ref{mismatchJ2_NG}, indicate that hybrid methods can effectively address system dynamics mismatch compared to MB methods. Furthermore, our TrackDiffuser outperforms KalmanNet.

\paragraph{Measurement rotation mismatch} 
Measurement rotation mismatch is simulated by generating data using a unit matrix with a slight rotation of \( \theta = 1^\circ \), while we use the unit matrix as the measurement matrix for training and testing. This rotation is equivalent to a sensor misalignment of approximately 0.55 \%. In this case, we again set \( \nu = -20 \, \text{[dB]} \) and \( T = 100 \). As shown in Figure \ref{mismatchRo_G} and \ref{mismatchRo_NG}, the results demonstrate that even this seemingly small rotation can lead to a significant performance degradation in MB filters. In contrast, the hybrid filters, KalmanNet and TrackDiffuser, can learn from the data to overcome this mismatch and significantly outperform MB methods. Furthermore, since TrackDiffuser does not require the measurement matrix, it exhibits superior performance and robustness compared to KalmanNet.

\subsubsection{Robustness under Training-Testing mismatch}

\begin{figure}[htb!]
    \centering
    \includegraphics[width=0.8\linewidth]{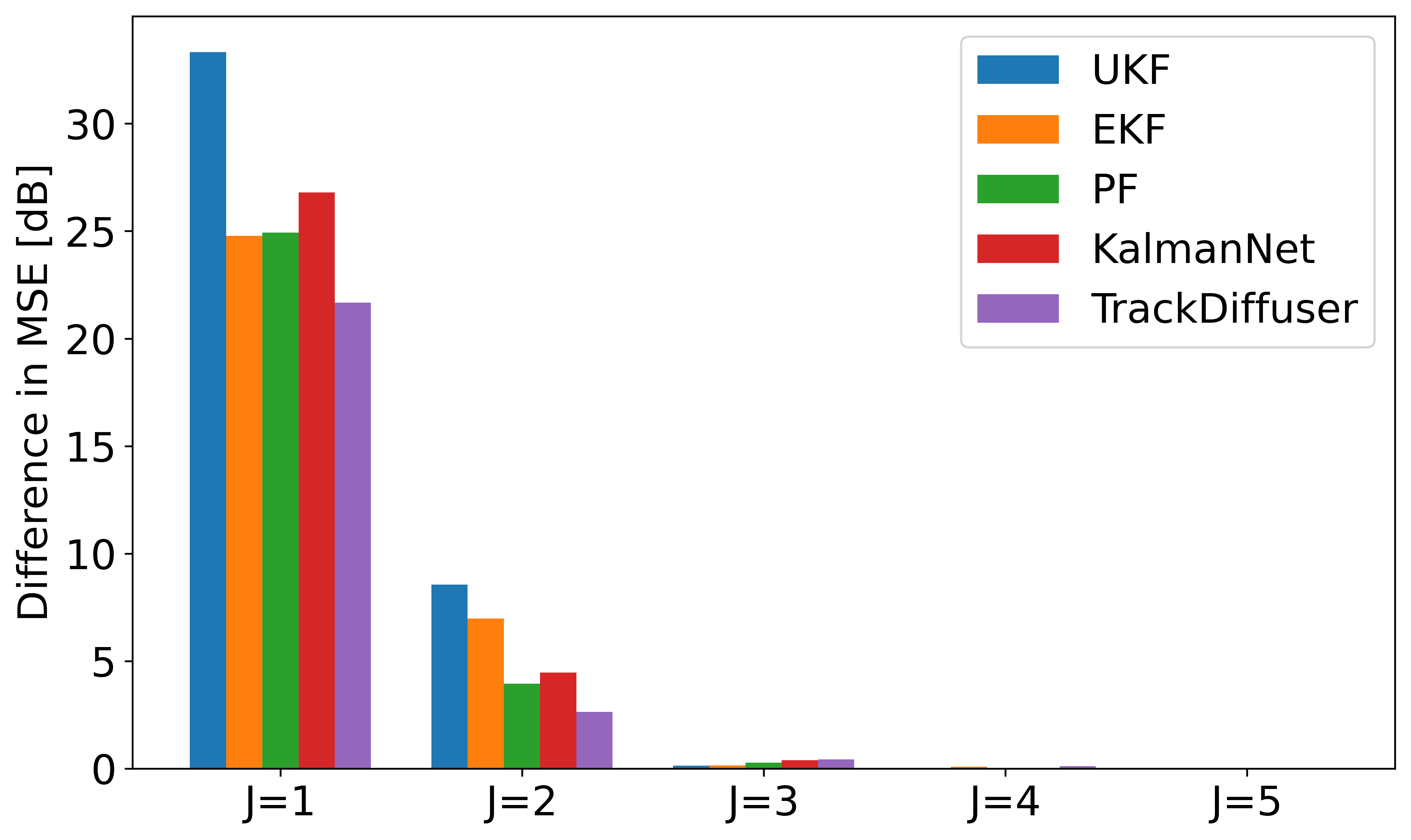}
    \caption{System dynamics training-testing mismatch. Performance degradation under approximation with $ J $ coefficients. Performance decline relative to the baseline with the accurate state-evolution function.}
    \label{fig9_FJ_diff}
\end{figure}

We aim to study the effectiveness in practical applications where the state-evolution function, approximated with $ J $ coefficients of Equation (\ref{6.0.2}), has not been used during training. 
In this case, we consider linear measurements with Gaussian noise, as outlined in Table \ref{table1}, with a fixed \(1/\mathrm{r}^2=20 \text{[dB]}\). 
KalmanNet and TrackDiffuser were trained on the accurate system dynamics model $\mathbf{f}(\cdot)$, while during testing, they performed filtering using the approximated state-evolution function with $ J=1\sim5 $. 
Figure \ref{fig9_FJ_diff} indicates that the model approximation has a significant impact on all methods. TrackDiffuser is the least affected and demonstrates superior robustness.

\subsection{Applications in real-world dynamics }

So far we have evaluated the performance of TrackDiffuser on the non-linear SSM - the Lorenz attractor. A natural question arises: how does it perform in real-world applications? As a result, we conducted an experiment using the Michigan NCLT dataset \cite{10.1177/0278364915614638}. In this case, our goal is to estimate the position of the Segway robot based on odometry measurements. 

\paragraph{Dataset processing} \label{Appendix_B}
This dataset includes labeled trajectory data of a moving Segway robot, where the measurements consist of odometry and GPS data. 
We utilized a dataset comprising a trajectory from the Segway robot date 2012-01-22. Due to unstable readings, we selected the first 20,000 data points and downsampled them at 5 Hz, yielding 1,000 time steps. These time steps were evenly divided into 25 trajectories, each containing 40 time steps. The trajectories were partitioned into three subsets: 17 trajectories for training, 3 for validation, and 3 for testing.

\paragraph{SSM for the Michigan NCLT dataset}
For the simulations on the Michigan NCLT Data Set, We define the state vector model as \( \mathbf{x_t} = (x_1, x_2, v_1, v_2)^{\top} \in \mathbb{R}^4 \), where \( x \) represent positions, and \( v \) represent velocities, the Segway SSM as a linear Wiener velocity model, where acceleration is modeled as a Gaussian white noise \cite{bar2004estimation}:

\begin{equation}\label{4.5.1}
    {x}_t = \mathbf{F}{x}_{t-1} + \mathbf{w}_t,
\end{equation} where \( \mathbf{F} \) is the motion matrix, process noise $ \mathbf{w}_t \sim \mathcal{N}(0, \mathbf{Q}) $ and

\begin{equation}\label{4.5.2}
\mathbf{F} = 
\begin{bmatrix}
1 & 0 & \Delta t & 0 \\
0 & 1 & 0 & \Delta t \\
0 & 0 & 1 & 0 \\
0 & 0 & 0 & 1
\end{bmatrix},
\quad
\mathbf{Q} = \text{q}^2
\begin{bmatrix}
\frac{\Delta t^3}{3} & 0 & \frac{\Delta t^2}{2} & 0 \\
0 & \frac{\Delta t^3}{3} & 0 & \frac{\Delta t^2}{2} \\
\frac{\Delta t^2}{2} & 0 & \Delta t & 0 \\
0 & \frac{\Delta t^2}{2} & 0 & \Delta t
\end{bmatrix},
\end{equation} where \( \Delta t \) represents the time interval between consecutive measurements. Since TrackDiffuser and KalmanNet, as hybrid methods, do not require prior knowledge of the noise covariance matrix \( \mathbf{Q} \). In this case, \( \mathbf{Q} \) is used during testing in MB methods.

In this case, the goal is to estimate the position of the Segway robot based on odometry data; i.e., measurements are given by processed odometry readings including velocity and computed relative position, and a measurement model as follows:
\begin{equation}\label{4.5.3}
    {z}_t = \mathbf{H}{x}_{t} + \mathbf{v}_t,
\end{equation} where measurement matrix $\mathbf{H} = \mathbf{I}_4$, measurement noise $ \mathbf{w}_t \sim \mathcal{N}(0, \mathbf{R})$ with $ \mathbf{R} = \mathrm{r}^2 \mathbf{I}_4 $.

It is important to note that since TrackDiffuser does not utilize $\mathbf{H}$, it requires full-dimensional state information. Consequently, the measurements need to be processed into computed relative positions and velocities, which contrasts with the direct use of velocity as the measurement vector. Furthermore, TrackDiffuser does not necessitate knowledge of process noise $\mathbf{w}_t$ or measurement noise $\mathbf{v}_t$. This allows TrackDiffuser to operate without accurate prior knowledge of SSM and directly estimating the posterior state based on the measurements. We compared the performance of TrackDiffuser, KalmanNet, and MB EKF.

\begin{figure}[htb!]
	\centering
	\includegraphics[width=0.8\linewidth]{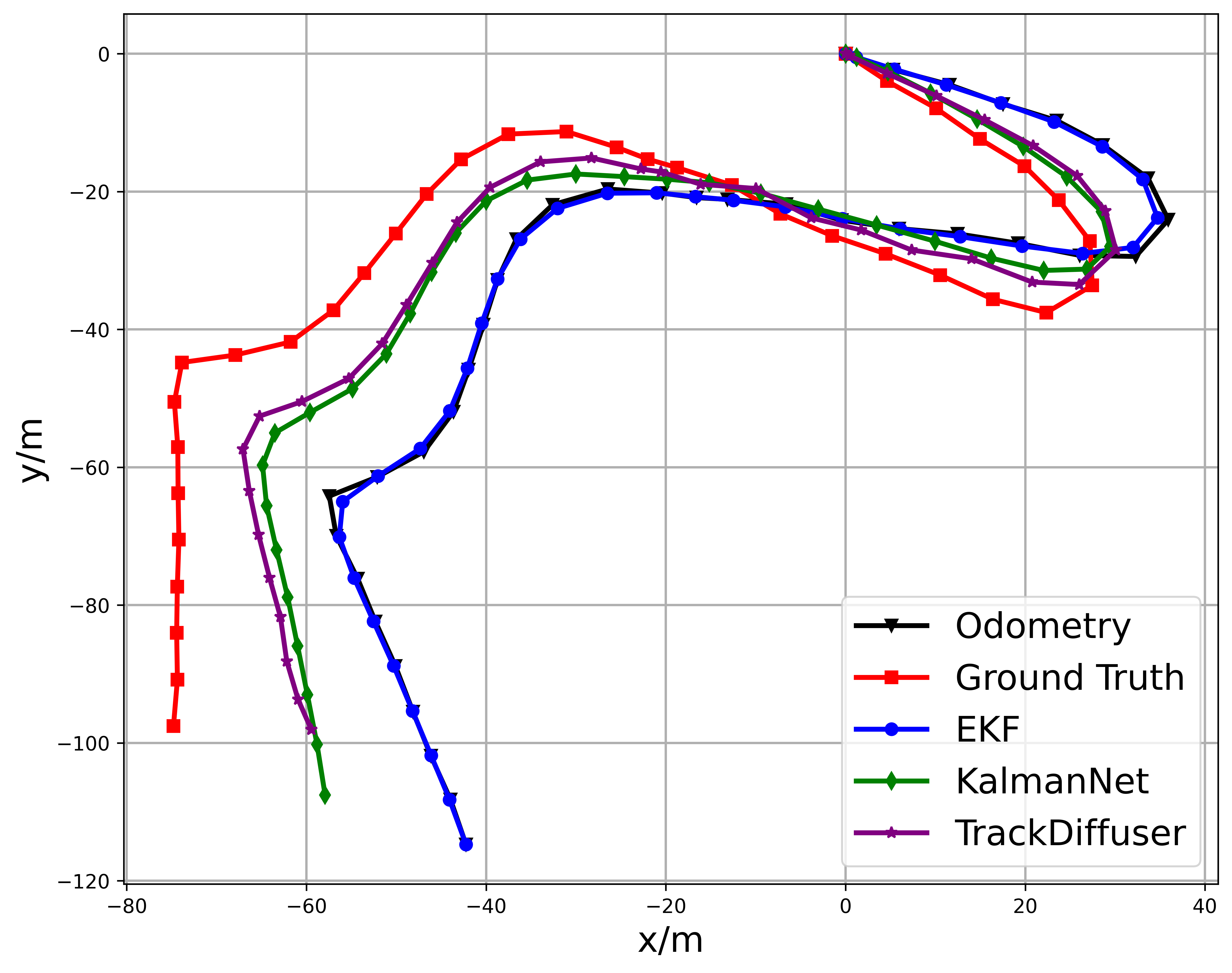} 
	\caption{NCLT data set, trajectory from
session with date 2012-01-22 sampled at 5 Hz.}
	\label{fig9}
\end{figure}

Table \ref{table9} and Figure \ref{fig9} illustrate the superiority of TrackDiffuser. 
MB filter is unable to address the measurement drift, leading to a similar estimation of the baseline derived from velocity integration. 
In contrast, the hybrid approache effectively mitigates the errors introduced by noisy odometry measurements, and TrackDiffuser provides better real-time position estimation compared to KalmanNet. 
This result highlights the benefits of integrating diffusion model with system dynamics model to compute the posterior distribution.

\begin{table}[h]
\centering
\caption{Numerical MSE [dB] for the NCLT experiment}
\label{table9}
\begin{tabular}{|c|c|c|c|}
\hline
 \textbf{Baseline} & \textbf{EKF} & \textbf{KalmanNet} & \textbf{TrackDiffuser} \\
\hline
20.11 & 19.02 &  16.01 & \textbf{13.94} \\
\hline
\end{tabular}%
\end{table}

\section{Conclusion}

This paper proposes a nearly model-free Bayesian filtering method with a diffusion model, TrackDiffuser, to address the state estimation problem, which implicitly learns the system dynamics model from trajectory data to mitigate the effects of inaccurate state space model. 
In the predict step, a system dynamics constraint module is incorporated to ensure interpretability. In the update step, the noise model is conditioned on the original measurements, simultaneously circumventing explicit measurement models and the noise priors. 
Our numerical study demonstrates that TrackDiffuser approximates the best posterior distribution in the dataset, and outperforms both classic model-based and hybrid methods. 
Additionally, it shows robustness under mismatch conditions.

%\bibliographystyle{alpha}
%\bibliography{ref}

\newcommand{\etalchar}[1]{$^{#1}$}

\end{document}